\documentclass[sigconf]{acmart}

\usepackage{graphicx}

\setcopyright{rightsretained}

\acmConference[ICVGIP'25]{16th Indian Conference on Computer Vision, Graphics and Image Processing}{December 17–19,2025}{Mandi, India}
\acmYear{2025}
\copyrightyear{2025}

\acmPrice{15.00}

\begin{document}
\title{Generating Key Postures of {\em Bharatanatyam} {\em Adavu}s  with Pose Estimation }

 \author{Jagadish Kashinath Kamble} 
 \authornote{Correspondence:jkkamble@pict.edu}
 \affiliation{%
   \institution{Indian Institute of Technology}
   \city{Kharagpur}
   \state{West Bengal}
   \country{INDIA}
   \postcode{721302} 
 }
  \email{jkkamble@pict.edu}

 \author{Jayanta Mukhopadhyay}
 \affiliation{%
   \institution{Indian Institute of Technology}
   \city{Kharagpur}
   \state{West Bengal}
   \country{INDIA}
   \postcode{721302}
 }
 \email{jay@cse.iitkgp.ac.in}

 \author{Debaditya Roy}
 \affiliation{%
   \institution{Indian Institute of Technology}
   \city{Kharagpur}
   \state{West Bengal}
   \country{INDIA}
   \postcode{721302}
 }
 \email{debaditya@cse.iitkgp.ac.in}  

 \author{Partha Pratim Das} 
 \affiliation{%
   \institution{Ashoka University}
   \city{Sonipat}
   \state{Haryana}
   \country{INDIA}
   \postcode{131029}
 }
 \email{partha.das@ashoka.edu.in}

\renewcommand{\shortauthors}{}

\begin{abstract}
Preserving intangible cultural dances—rooted in centuries of tradition and governed by strict structural and symbolic rules—presents unique challenges in the digital era. Among these, Bharatanatyam, a classical Indian dance form, stands out for its emphasis on codified adavus and precise key postures. Accurately generating these postures is crucial not only for maintaining the dance’s anatomical and stylistic integrity, but also for enabling its effective documentation, analysis, and transmission to broader, global audiences through digital means. We propose a pose-aware generative framework integrated with a pose estimation module, guided by keypoint-based loss and pose consistency constraints. These supervisory signals ensure anatomical accuracy and stylistic integrity in the synthesized outputs. We evaluate four configurations: standard Conditional Generative Adversarial Network (cGAN), cGAN with pose supervision, conditional diffusion, and conditional diffusion with pose supervision. Each model is conditioned on key posture class labels and optimized to maintain geometric structure. In both cGAN and conditional diffusion settings, the integrated pose guidance aligns generated poses with ground-truth keypoint structures, promoting cultural fidelity. Our results demonstrate that incorporating pose supervision significantly enhances the quality, realism, and authenticity of generated {\em Bharatanatyam} postures. This framework provides a scalable approach for the digital preservation, education, and dissemination of traditional dance forms—enabling high-fidelity generation without compromising cultural precision. Code is available at \url{https://github.com/jagidsh/Generating-Key-Postures-of-Bharatanatyam-Adavus-with-Pose-Estimation}.

\end{abstract}

%



\maketitle

\section{Introduction}
Among the various forms of cultural practices, classical and traditional dances occupy a central role as part of humanity’s Intangible Cultural Heritage (ICH), as recognized by UNESCO ~\cite{unesco}. 
These dance forms are not merely artistic performances but codified systems of movement, gesture, and rhythm that convey deep cultural narratives, religious symbolism, and collective identity. 
Preserving and studying such dances is essential not only for cultural continuity but also for understanding the nuanced ways in which tradition is embodied and transmitted. 
India is home to a rich spectrum of classical dance  forms which are institutionalized but remain largely inaccessible due to their complexity \cite{Dhananjayan2002Transmission,Vatsyayan1997Theatre,ChallengesClassical2016,ClassicalGatekeeping2021}.
Despite institutional efforts, mastering these dance forms often requires years of guided training under expert supervision—something that is not always feasible for learners from different geographies, socioeconomic backgrounds, or non-traditional settings. The symbolic and structural intricacy of these dances presents significant barriers to access, especially in the absence of personalized guidance. In this context, AI-based generation methods offer a transformative opportunity. 
\begin{figure*}[htbp]
    \centering
    \includegraphics[width=\linewidth, height=5cm]{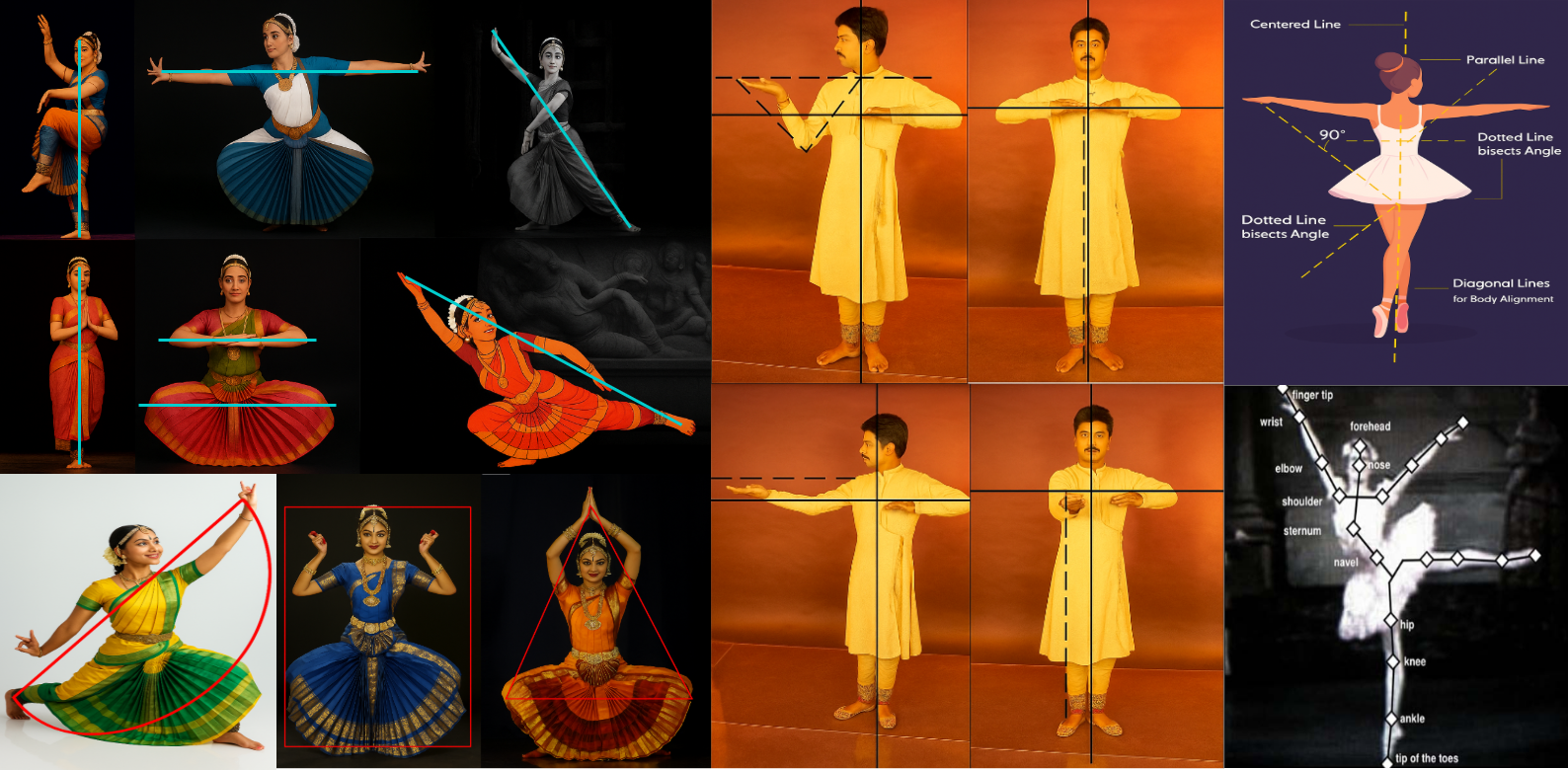}
    \caption{Visual comparison of posture alignment across classical dance forms. a: {\em Bharatanatyam} postures illustrating vertical, horizontal, and diagonal axes and body symmetry (source \cite{GeometryInMotion2021}). b: {\em Kathak} poses emphasizing upright torso and mirrored hand gestures (source \cite{Mallick2017KathakGeometry}). c: {\em Ballet} positions such as arabesque and plié, showing turnout and axial balance (source \cite{Daprati2009DanceToTheMusic,BalletBasics2015}). Despite cultural differences, these styles share a common reliance on geometrically structured key postures, justifying alignment-preserving approaches in generative modeling.}
    \label{fig:global_alignment}
\end{figure*}

\begin{figure*}
  \centering
 \includegraphics[width=0.9\textwidth]{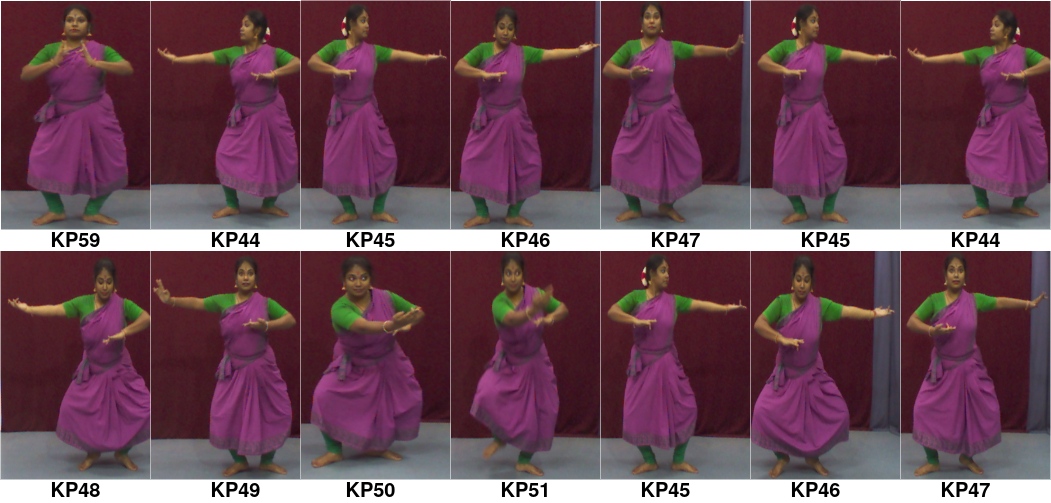}
    \caption{Illustration of Kuditta Tattal \textit{Adavu} sequence in Bharatanatyam showcasing the structured progression of body postures. Each frame represents a temporally ordered key pose, reflecting the codified nature of movement transitions in the dance form.}
\label{sample_kp}
\end{figure*}

\textit{Bharatanatyam}, one of the eight officially recognized forms of \textit{Indian Classical Dance} (ICD), is rooted in a codified system of symbolic gestures, narrative expression, geometric precision~\cite{kalpana2015bharatanatyam,ranganathan_design_bharatanatyam}, intricate grammar and choreographic discipline. Classical dance forms across the world—such as Indian Kathak dance, Chinese classical dance, Japanese Noh, Balinese dance, and European ballet—share a deep reliance on structured body geometry, posture alignment, and movement symmetry. 
These dance forms, while culturally distinct, are united by their emphasis on codified movement vocabularies and compositional rules make it ideal for precise pose generation\cite{Vatsyayan1997Theatre}. As illustrated in Figure~\ref{fig:global_alignment}, forms like \textit{Bharatanatyam}, \textit{Kathak}, and \textit{Ballet} all exhibit a shared reliance on posture precision, geometrical structure, and symmetrical alignment, which makes them suitable candidates for computational modeling. For instance, in ballet \cite{Morris01042003}, notes that strict stylistic codification can sometimes obscure interpretive depth, especially when movement is taught devoid of its expressive or contextual grounding. A parallel issue exists in Bharatanatyam, where the richness of \textit{Adavu} postures depends on nuanced body alignment, rhythm synchronization, and symbolic expression—elements that are difficult to preserve or analyze without expert intervention. In \textit{Bharatanatyam}, fundamental unit of movement, the \textit{Adavu},consists of a sequence of \textit{Key Postures} (KPs) (refer Figure~\ref{sample_kp}), which function as structural anchors that govern motion, rhythm, and visual balance. These key postures are not arbitrary—they conform to principles of \textit{angashuddhi} (correct body alignment) and \textit{nritta} (pure movement). A complete Adavu includes an initial pose, intermediate transitions, and a final posture, all choreographed to maintain harmony between the limbs, torso, head, and gaze while aligning with rhythmic syllables known as \textit{Sollukattu}. These KPs are foundational to both static and transitional elements of the performance. Although full posture sequences (refer Figure~\ref{sample_kp}) are visualized to demonstrate the formation of adavus, textual explanations of individual postures remain challenging due to the structural richness, interdependency among body parts that define each pose, symbolic, geometric, and rhythmic complexity inherent in Bharatanatyam. Every Bharatanatyam dance posture embodies a synthesis of discipline, symbolism, and aesthetic control. These poses are not merely biomechanical configurations but culturally codified units of expression, often associated with emotional states, narratives, or devotional gestures. 

Synthesizing Bharatanatyam key postures is therefore central to generating Adavu sequences. Unlike contemporary dance, which often permits freeform or improvised motion, Bharatanatyam requires exact control and coordination of the body through codified poses. These key postures define not only anatomical structure but also expressive content. Their accurate reproduction—particularly when synchronized with rhythmic syllables—preserves the dance's intended rhythm, clarity, and visual coherence, enabling it to be analyzed, taught, or reconstructed digitally. Such synthesis is vital for dance education, cultural preservation, and broader computational understanding of classical dance grammar. While grounded in Bharatanatyam, our approach extends to other classical dance forms such as Kathak and Ballet, which also rely on symmetry, alignment, and structured body mechanics. Figure~\ref{fig:global_alignment} illustrates this shared emphasis on geometric precision across styles, underscoring the global significance of posture-aware generative modeling. A curated library of synthesized poses supports realistic Adavu generation, smooth temporal interpolation, motion analysis, and virtual choreography. 

\subsection{Challenges}
\label{sec:challenges}
We identify various challenges for the tasks at hand, including:
\begin{itemize}
    \item \textbf{Spatial Accuracy and Classical Complexity:} Classical dance key postures demand precise spatial relationships between body parts to maintain anatomical symmetry and cultural authenticity. Unlike general pose generation, these poses follow strict rules involving complex limb coordination and symbolic gestures, where even minor misalignments can distort their traditional meaning.

     \item \textbf{Scarcity of Domain-Specific Key Posture Datasets:} A key bottleneck is the absence of high-quality datasets specifically curated for classical dance postures. While many pose datasets focus on contemporary actions or dance genres (e.g., AIST++ or NTU RGB+D)~\cite{li2021learn, NTU_RGBD}, they lack the annotated precision and cultural granularity needed for traditional forms like Bharatanatyam. Building and using such a dataset requires expert input to accurately label mudras, body symmetry, and pose correctness according to classical norms.
\end{itemize}
To address these challenges, we design a system that automatically generates classical dance key postures based on specific labels (like different keyposture types). To ensure the generated postures are anatomically correct and culturally accurate, we include a special pose-aware guidance mechanism to check body symmetry and structure using two losses—keypoint loss and pose consistency loss—which help the model keep the dancer's posture and gestures aligned with classical rules.

Our Key Contributions are as follows:
\begin{itemize}
   
\item We present the first structured attempt at generating anatomically and culturally faithful Bharatanatyam key postures using deep generative models, addressing strict spatial, symbolic, and aesthetic constraints of classical Indian dance—a domain overlooked in prior pose generation research.

\item We incorporate pose-aware guidance mechanism into the generative pipeline to enforce keypoint-level and pose-level consistency losses, ensuring accurate skeletal alignment, joint symmetry, and relative body part placement across frames.

\item We perform a comprehensive comparison between different generative models with and without pose estimation—to analyze their performance on structure-aware human image synthesis in a culturally constrained context. Our evaluation includes FID and MS-SSIM metrics computation along with detailed visual comparisons. Results show that the Conditional Diffusion with Pose Estimation model best preserves stylistic fidelity and structural accuracy.
\end{itemize}

\begin{figure*}[ht]
    \centering
    \includegraphics[width=\linewidth]{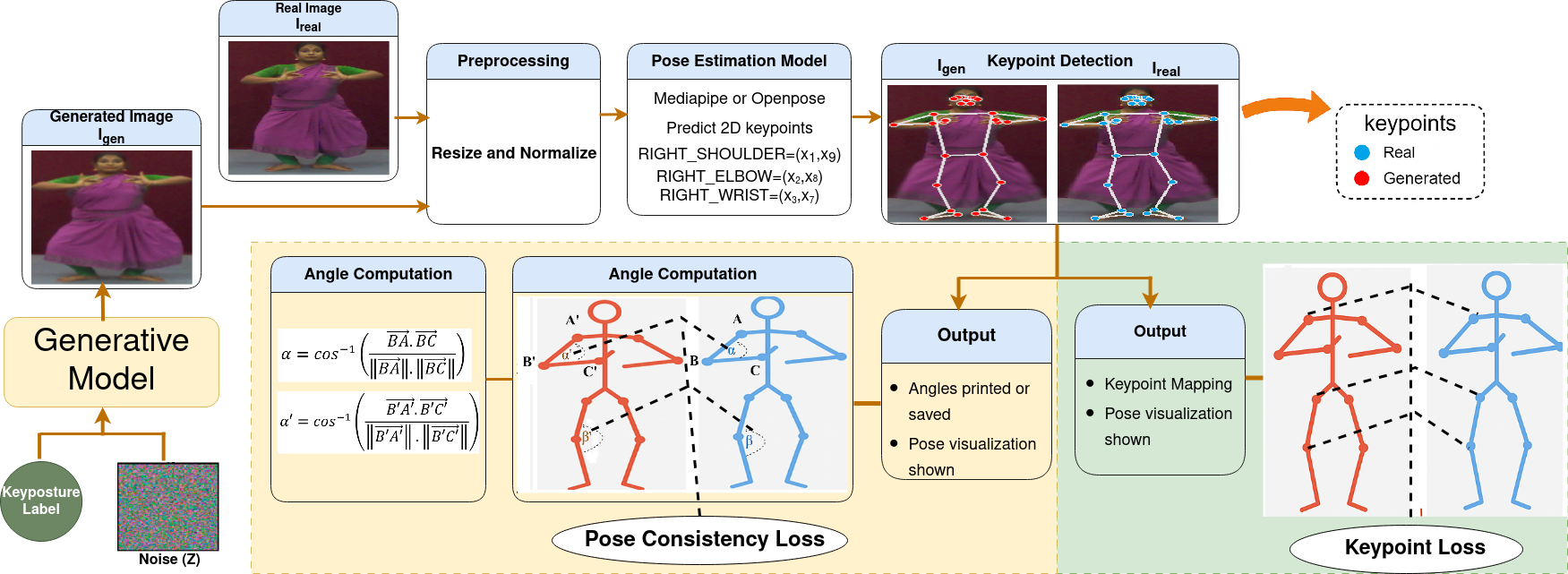}
    \caption{Overview of the proposed model. A conditional generative model generates a dance image \( I_{\text{gen}} \) from a latent code \( z \) and a key posture label. Pose Estimation Module is used to extract 2D skeletal keypoints.  \textit{Keypoint Loss} -- (\( \mathcal{L}_{\text{kp}} \)) average squared Euclidean distance between normalized corresponding keypoints from \( I_{\text{gen}} \) and \( I_{\text{real}} \), enforcing pixel-wise accuracy of each joint. \textit{Pose Consistency Loss} -- (\( \mathcal{L}_{\text{pose}} \)) difference in internal pose structure capturing relative distances and angles between joints—between the generated and real poses.
       }
    \label{fig:architecture}
\end{figure*}
\section{Related work}
\subsection{Human Image Synthesis}

Human image synthesis has become a central topic in computer vision, with increasing demand for controllable and realistic generation of human appearances. Applications span across entertainment, virtual try-on, telepresence, and cultural preservation. Early progress was driven by the success of \textit{Generative Adversarial Networks (GANs)\cite{mirza2014conditional,goodfellow2014generative}}, which laid the foundation for generating photo-realistic human images from latent spaces. Traditional methods treated humans as monolithic objects and struggled with the diversity of human poses, appearances, and clothing. \cite{Fruehstueck2022InsetGAN} introduce InsetGAN, a method for full-body image generation that conditions on local image patches to preserve fine details while maintaining global coherence. It enables controllable synthesis of human images, showing strong results on fashion datasets like DeepFashion.

\subsection{Pose-Guided and Skeleton-Based Synthesis}
Pose-guided synthesis—also referred to as \textit{skeleton-based person generation}—focuses on conditioning human image generation on pose representations, typically given as 2D/3D skeleton keypoints or heatmaps. This subfield enhances control and realism, making it particularly useful in scenarios like virtual avatar creation, action retargeting, and artistic motion generation.Foundational works like Pose-Guided Person Image Generation (PG2)~\cite{ma2017pose} and its disentangled variant~\cite{ma2018disentangled} pioneered this direction by introducing two-stage architectures to extract appearance from a source image and render it under a target pose. These architectures demonstrated the feasibility of generating structurally plausible human images from pose inputs. \cite{Liang2019PCGAN} propose a two-stage framework for pose-guided person image generation. Given a reference image and target pose, the model first synthesizes a coarse output via pose integration, then refines it using adversarial training to produce realistic person images in new poses. \cite{lee2023embedding} introduce a diffusion-based framework for pose-guided person image synthesis, leveraging embedding fusion to integrate pose and appearance cues. The method enhances structural consistency and image quality, outperforming traditional GAN-based approaches. Unsupervised Parsing-Guided GAN~\cite{song2019unsupervised} and Look into Person (LIP-GAN)~\cite{gong2017look} decomposed the human body into semantic parts (e.g., arms, legs, torso), which helped guide localized transformations and ensure that synthesized parts were semantically aligned with target poses. \cite{zhao2022stylegan} present StyleGAN-Human, a high-quality human image generator trained on a curated dataset of full-body human images. By integrating pose and appearance controls into the StyleGAN architecture, the model enables fine-grained synthesis and editing of realistic human images with controllable structure and style. 
\subsection{Dance Posture Synthesis and Movement Modeling} 

While Bharatanatyam is a highly codified and symbolically rich classical dance form, there is a surprising lack of research focused directly on the automatic synthesis of its key postures. Most prior work in Indian Classical Dance (ICD) research has concentrated on pose recognition, gesture (mudra) classification, motion capture, and pose estimation tasks using traditional computer vision or learning-based approaches~\cite{bhuyan2021keyframes, kaushik2018nrityantar, mallick2019posture, raju2024pose2gest, jain2021enhanced, kishore2018indian, kumar2017indian, jayanthi2024ai}. However, the generative modeling of classical dance movements — particularly Bharatanatyam's nuanced postures and transitions — remains underexplored. Consequently, we draw upon parallel research areas such as data-driven human motion generation, pose-conditioned synthesis using GANs, diffusion models, and transformers. For example, works like \cite{chan2019everybody} and \cite{wang2024musicdrivendance} emphasize the role of music-aligned movement generation and spatial-temporal modeling to produce fluid dance sequences. The POPDG dataset introduced by \cite{luo2024popdg} further showcases the need for genre-specific datasets, underscoring how aesthetics and cultural context influence the effectiveness of generative frameworks. Moreover, \cite{seong2021keypoint} demonstrate how loss functions tuned to keypoint-specific errors can significantly improve structural realism in synthesized poses — an insight critical for preserving the formal grammar of Bharatanatyam. Despite these advances, most existing works focus on generic street scenes or fashion data~\cite{liu2016deepfashion, jetchev2017conditional}, with limited attention to culturally rich, structured movements as in classical dance. 

Our work differentiates itself by tailoring pose synthesis to domain-specific constraints of classical dance, which involves stringent spatial alignment, symbolic gestures, and rhythm-aware transitions—challenges often overlooked in prior research. By integrating fine-grained pose estimation~\cite{cao2017realtime, balakrishnan2018synthesizing} and generative modeling~\cite{ho2020denoising,mirza2014conditional}, we aim to bridge this gap and preserve the stylistic fidelity of culturally significant dance forms.

\section{Methodology}
\label{sec:methodology}

\subsection{Problem Formulation}

We address the task of synthesizing realistic and anatomically correct key postures for \textit{Bharatanatyam adavus}—the foundational movement units in classical Indian dance. We formulate this as a conditional image generation problem, wherein the objective is to learn a function:
\[
G: (z, y) \rightarrow I
\]
Here, \( z \sim \mathcal{N}(0, I) \) denotes a Gaussian noise vector capturing stochastic variation, \( y \in \mathcal{Y} \) is the class label indicating the target \textit{adavu} category, and \( I \in \mathbb{R}^{H \times W \times 3} \) is the generated RGB image representing the dance posture.

Conditional generative frameworks such as Conditional GANs~\cite{mirza2014conditional} or Conditional Diffusion Models~\cite{ho2020denoising,ho2022classifier} can synthesize visually plausible outputs from these conditional inputs. However, these models often introduce distortions in body orientation and joint alignment~\cite{deBem2020DGPose,Xu2023PACGAN,Zhang2024MimicMotion} due to their inherent bias toward diversity. Such variability, while desirable in generic generation tasks, is unsuitable in the context of classical dance, where postural precision and adherence to culturally defined constraints are non-negotiable.

To address this, we propose a \textbf{pose-aware guidance mechanism} integrated into the generative pipeline. This mechanism supervises the generation process to ensure that the synthesized dance postures preserve rule-based anatomical fidelity and alignment.

\subsection{Pose-Aware Supervision via Joint Keypoint Constraints}
\label{pose_model}

Classical dance forms like \textit{Bharatanatyam} are characterized by highly codified postural geometries~\cite{kalpana2015bharatanatyam,ranganathan_design_bharatanatyam,GeometryInMotion2021,Mallick2017KathakGeometry,Daprati2009DanceToTheMusic,BalletBasics2015} and symbolic expressions~\cite{lokumannage2021bharatanatyam,sharma2013bharatanatyam,chakraborty2019dance}. Key positions such as \textit{Araimandi} (symbolizing humility) or \textit{Tribhangi} (evoking divine energy) demand precise limb configurations, angular alignments, and symmetry (see Figure~\ref{fig:global_alignment}).

To enforce these structural constraints, we incorporate a pre-trained pose estimation module (e.g., \textit{MediaPipe}~\cite{mediapipe2020}) into the training process. This module extracts body joint keypoints from both real and generated images, enabling structural supervision. The predicted keypoints are denoted by:
\begin{equation}
P = \{(x_i, y_i)\}_{i=1}^{K}
\label{eq:kp}
\end{equation}
where $K$ is the total number of body joints, and $(x_i, y_i)$ are the 2D coordinates of the $i$-th joint.

Let the generator produce a synthetic image $\hat{I} = G(z, y)$ and $I$ denote its ground-truth counterpart. The pose estimation model $E(\cdot)$ extracts keypoints from both:
\[
\hat{P} = E(\hat{I}) \in \mathbb{R}^{K \times d}, \quad P = E(I) \in \mathbb{R}^{K \times d}
\]
where $d$ is typically 2 or 3 (for 2D or 3D keypoints). The extracted keypoints guide the generation using two loss functions described below.

\subsubsection{Keypoint Alignment Loss}
\label{subsec:kp_loss}

The \textbf{Keypoint Alignment Loss} ($\mathcal{L}_{\text{kp}}$) enforces pixel-level accuracy of joint placement. It penalizes deviations in the normalized keypoint coordinates between generated and ground-truth images:
\begin{equation}
\mathcal{L}_{\text{kp}} = \frac{1}{K} \sum_{i=1}^{K} 
\left\| 
\text{Norm}((x_i^{\text{gen}}, y_i^{\text{gen}})) - 
\text{Norm}((x_i^{\text{gt}}, y_i^{\text{gt}})) 
\right\|^2
\label{eq:kp_loss}
\end{equation}
where $\text{Norm}(\cdot)$ denotes a normalization function (e.g., scale normalization based on torso length or image dimensions). This loss ensures anatomical precision by aligning individual joints, critical for modeling structured dance postures.

\subsubsection{Pose Consistency Loss}
\label{subsec:pose_consistency}

The \textbf{Pose Consistency Loss} ($\mathcal{L}_{\text{pose}}$) focuses on preserving the internal geometric relationships among joints. Rather than enforcing absolute coordinates, it evaluates relative distances and angles between joints to encourage structure-aware generation. Formally:
\begin{equation}
    \mathcal{L}_{\text{pose}} = \sum_{j=1}^{N} \left( R^{\text{gen}}_j - R^{\text{gt}}_j \right)^2
\end{equation}
where $R_j^{\text{gen}}$ and $R_j^{\text{gt}}$ are the $j$-th relative pose feature (e.g., joint-to-joint distance or angular orientation) computed from the generated and ground-truth poses respectively, and $N$ is the total number of such features.

For example, given the left shoulder, elbow, and wrist positions, the distance and angle formed by these points in the generated image must match that in the real image. This loss is invariant to global translation or rotation and promotes symmetry and geometric fidelity, both of which are essential in classical dance synthesis.

\subsection{Pose-Aware Generative Framework}
\label{subsec:pose_aware_framework}

We now present our general framework for pose-aware conditional image generation. This framework is designed to be model-agnostic and can be integrated with any conditional generative architecture (e.g., GANs, VAEs, or diffusion models).

\paragraph{Generation:} Given a class label $y \in \mathcal{Y}$ and a latent noise vector $z \sim \mathcal{N}(0, I)$, the generator synthesizes an image $\hat{I} = G(z, y)$. A discriminator (in GANs) or noise scheduler (in diffusion models) supervises visual plausibility.

\paragraph{Pose-aware Supervision:} To ensure postural correctness, a pose estimation network $E(\cdot)$ extracts skeletal keypoints from both $\hat{I}$ and the corresponding real image $I$. These keypoints are used to compute the $\mathcal{L}_{\text{kp}}$ and $\mathcal{L}_{\text{pose}}$ losses as defined above.

\paragraph{Total Objective:} The generator is optimized using a composite loss:
\begin{equation}
\mathcal{L}_{\text{total}} = \mathcal{L}_{\text{gen}} + \mathcal{L}_{\text{kp}} + \lambda_{\text{pose}} \mathcal{L}_{\text{pose}}
\end{equation}
Here, $\mathcal{L}_{\text{gen}}$ represents the model-specific generation loss (e.g., adversarial loss $\mathcal{L}_{\text{adv}}$ for GANs or conditional reconstruction loss $\mathcal{L}_{\text{recon}}$ for conditional diffusion models), and $\lambda_{\text{pose}}$ controls the strength of pose-level supervision. In our experiments, we set $\lambda_{\text{pose}} = 1$ for effective yet stable training.

\paragraph{Modularity and Integration:} As illustrated in Figure~\ref{fig:architecture}, the proposed architecture seamlessly incorporates the pose-aware module into any generative pipeline. By aligning the generation process not only with visual features but also with structural pose features derived from classical form, our framework ensures the synthesis of visually realistic, geometrically accurate, and semantically faithful dance postures.

\section {Experimental Results and Discussion}
We conducted a comprehensive evaluation of our proposed approach, which integrates Conditional Generative Adversarial Networks (cGANs) and Conditional Diffusion models with or without a pose estimation module, to synthesize visually realistic and structurally accurate Bharatanatyam postures. The system is specifically tailored for a curated Bharatanatyam dataset\cite{bhuyan2021keyframes,mallick2019posture} and employs MediaPipe-based keypoint loss and pose consistency loss to enhance the generated outputs by aligning them closely with real human pose structures and the codified symmetry found in classical dance.

\subsection{Implementation Details}

We use two prominent image generation models -- conditional GANs (CGAN) and conditional diffusion, to test our pose estimation module.

(a) \textit{Conditional GAN} \cite{mirza2014conditional} model is trained with a batch size of 10, a learning rate of 0.0002, and the Adam optimizer with the following hyperparameters: \(\beta_1 = 0.5 \quad \text{and} \quad \beta_2 = 0.999\). 
The generator of the conditional GAN (cGAN) takes a latent noise vector \( z \) and a key posture label, which is first converted into a continuous embedding via an embedding layer. These are concatenated and passed through a series of fully connected layers to synthesize key posture images. The discriminator receives both the generated or real image and the corresponding label embedding to jointly evaluate whether the image is real or fake.

To enforce anatomical correctness, we integrated our pose estimation module based on MediaPipe to extract 2D human keypoints from both real and generated images. 
We integrate \textit{keypoint loss} (\( \mathcal{L}_{\text{kp}} \)) and \textit{pose consistency loss} (\( \mathcal{L}_{\text{pose}} \)) into the training objective of CGAN as 



\[
\mathcal{L}_{\text{total}} = \mathcal{L}_{\text{adv}} + \lambda_{\text{kp}} \mathcal{L}_{\text{kp}} + \lambda_{\text{pose}} \mathcal{L}_{\text{pose}}
\]

where \( \mathcal{L}_{\text{adv}} \) is the adversarial loss of the CGAN, \( \mathcal{L}_{\text{kp}} \) minimizes the distance between the keypoints of real and generated images, and \( \mathcal{L}_{\text{pose}} \) ensures global pose consistency.

(b) \textit{Conditional Diffusion model}~\cite{ho2020denoising,ho2022classifier} comprises of an U-Net backbone. The model consists of four hierarchical encoding and decoding blocks, with a bottleneck residual block connecting the two. 
Each encoder stage includes two residual blocks with convolutional downsampling, and each decoder performs 2$\times$ upsampling using nearest-neighbor interpolation followed by convolution. Skip connections preserve spatial resolution across the encoder-decoder path. This model is conditioned on \textit{sinusoidal time embeddings} representing the diffusion timestep and \textit{class embeddings} for label-aware generation. Additionally, a \textit{self-attention module} is introduced at select resolutions to capture long-range spatial dependencies critical for the structural fidelity of pose generation.

The \textit{keypoint loss} and \textit{pose consistency loss} are incorporated into the diffusion training objective as follows:

\[
\mathcal{L}_{\text{total}} = \mathcal{L}_{\text{recon}} + \lambda_{\text{kp}} \mathcal{L}_{\text{kp}} + \lambda_{\text{pose}} \mathcal{L}_{\text{pose}}
\]
Here, \(\mathcal{L}_{\text{recon}}\) is the conditional reconstruction loss computed across all time-steps in the diffusion process:

\[
\mathcal{L}_{\text{recon}} = \mathbb{E}_{x_t, y, t, \epsilon} \left[ \left\| \epsilon - \epsilon_\theta(x_t, y, t) \right\|_2^2 \right]
\]

where \(x_t\) is the noisy image at diffusion timestep \(t\),\(y\) is the conditioning input (e.g., class label or pose),\(\epsilon\) is the true noise added to the clean image,\(\epsilon_\theta(x_t, y, t)\) is the noise predicted by the model,\(\| \cdot \|_2^2\) denotes the mean squared error, while \( \mathcal{L}_{\text{kp}} \) and \( \mathcal{L}_{\text{pose}} \) function as auxiliary constraints to guide the model toward anatomically valid outputs.

\begin{figure}[h]
  \centering
  \includegraphics[width=\linewidth]{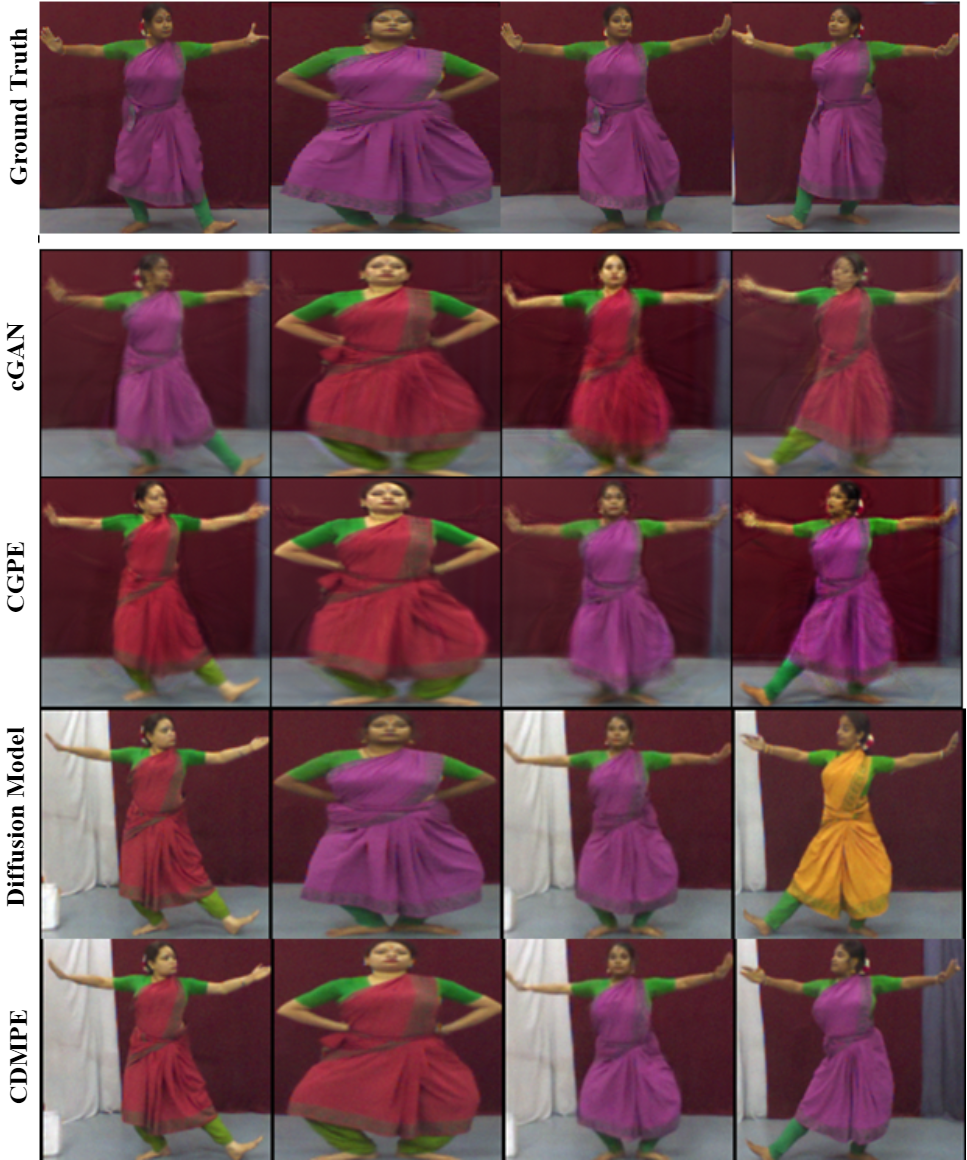}
  \caption{Comparison of generated Adavu key postures across four models—(a) Conditional GAN (cGAN), (b) cGAN with Pose Estimation (CGPE), (c) Conditional Diffusion, and (d) Conditional Diffusion with Pose Estimation (CDMPE). Real reference images (ground truth) are shown alongside for comparison. While all models produce visually plausible results, the diffusion-based models demonstrate smoother textures and higher fidelity. The Conditional Diffusion with Pose Estimation (d) exhibits the most anatomically accurate and culturally faithful postures, preserving structural constraints and symbolic hand-gesture alignment essential to Bharatanatyam.}
\label{result}
\end{figure}
\begin{figure*}[htb]
  \centering
  \includegraphics[width=0.99\textwidth, height=0.25\textheight]{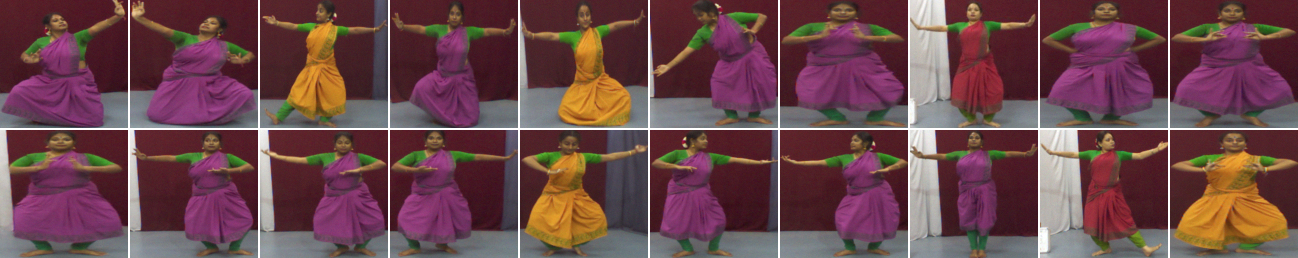}
 \caption{Complete set of generated Adavu key postures ($\mathbf{128 \times 128}$) using the Conditional Diffusion with Pose Estimation model. All 20 representative postures are synthesized to reflect structural precision, anatomical symmetry, and cultural authenticity specific to Bharatanatyam. This visualization highlights the model’s ability to generalize across a diverse set of codified movements while preserving stylistic consistency.}
\label{best_result}
\end{figure*} 
\subsection{Dataset Preparation}
\label{sec:dataset}
For the synthesis and analysis of {\em Bharatanatyam} dance movements, we curated a comprehensive data set \cite{bhuyan2021keyframes,mallick2019posture} specifically tailored to capture the intricate nuances of this classical Indian dance form. The data set consists of recordings from 10 dancers performing 15 foundational {\em Adavu} with a total of 58 distinct variants. These performances were meticulously annotated to ensure precise characterization and provide a rich resource for Dance analysis and generation. The dataset comprises 1,276 high-quality videos, capturing the dancers' movements across various angles and conditions to ensure diversity and robustness. Furthermore, 334 unique motion sequences have been identified and categorized, enabling a comprehensive understanding of the dynamic transitions between poses. This data set, which combines visual, auditory, and motion data, provides a foundational resource for understanding and synthesizing key postures and {\em adavu} movements. 

\subsection{Quantitative Evaluation} Our task requires generating identical key postures within each class, leading to low intra-class diversity. To evaluate model performance, we computed {\em Fréchet Inception Distance} (FID) \cite{heusel2017gans}  using a pre-trained Inception V3 classifier. We fine-tuned\cite{liu2022improved} it on our dataset to better ~\cite{jayasumana2024rethinkingfidbetterevaluation} capture pose geometry, hand gestures ({\em mudra}s), and cultural nuances, improving FID reliability. We also computed multi-scale structural similarity index (MS-SSIM) \cite{wang2003multiscale} which measures the structural similarity between generated and real images at multiple scales. The quantitative results of our models, evaluated using FID Score and MS-SSIM are presented in Table~\ref{tab:pose_module_impact}, which demonstrates the impact of pose estimation modules on synthesis quality. Furthermore, the ablation study highlighting the influence of different loss components is detailed in Table~\ref{tab:loss_impact}.
Since FID inherently favors datasets with higher intra-class variation\cite{lucic2018gans,sajjadi2018assessing,borji2019pros}, our results show a saturation point in FID improvement. This is not due to poor generation quality but rather due to the nature of key posture consistency and repetitive key posture generation. Our results show a MS-SSIM score consistently above 0.74, indicating good structural similarity with real images, indicative of good perceptual quality, reflecting well preserved spatial relationships and textures \cite{wang2003multiscale, borji2019pros}. 

\begin{figure}[h]
    \centering
 \includegraphics[width=\linewidth]{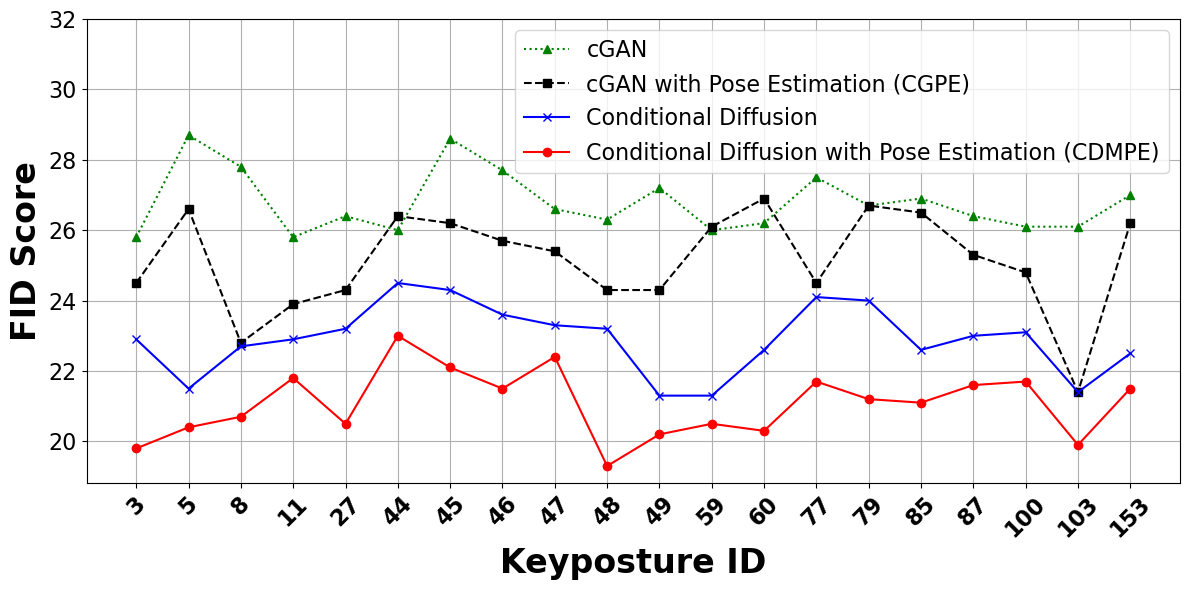}
\caption{FID score comparison for individual Bharatanatyam key postures across four generation models—cGAN, cGAN with pose estimation, cDiffusion, and cDiffusion with pose estimation. Lower FID(↓) values indicate better generation fidelity.}
    \label{fig:fid_keyposture_metrics}
\end{figure}

\begin{figure}[h]
    \centering
 \includegraphics[width=\linewidth]{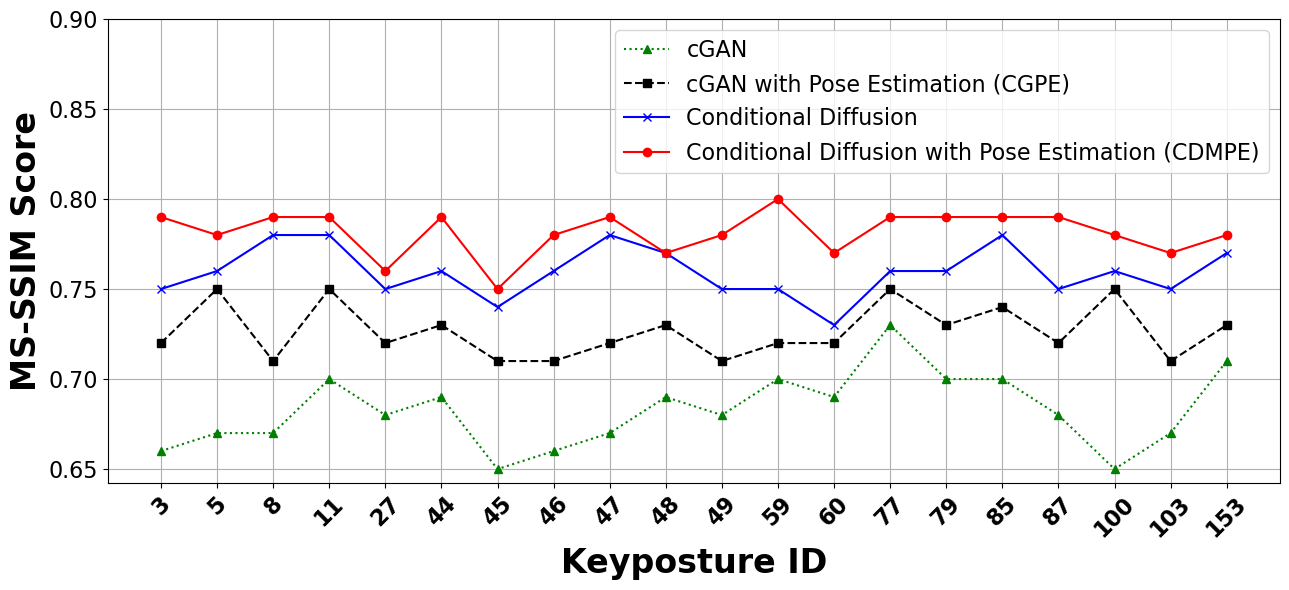}
\caption{MS-SSIM scores for Bharatanatyam key postures across various generative models-—cGAN, cGAN with pose estimation, cDiffusion, and cDiffusion with pose estimation. Higher values (↑) represent better structural similarity and perceptual quality.}
    \label{fig:msssim_keyposture_metrics}
\end{figure}

The structural quality of the generated key postures is promising, particularly in maintaining body pose coherence, as evidenced by the qualitative results presented in Figure ~\ref{result} and ~\ref{best_result}. The model effectively captures the spatial relationships of key joints, making notable progress in representing dance movements, which will be evident in the visual representation of the Key posture.
To comprehensively evaluate the performance of all models, we present a keyposture-wise comparison across the metrics—FID and MS-SSIM—for 20 representative keypostures (Figure~\ref{fig:fid_keyposture_metrics} and ~\ref {fig:msssim_keyposture_metrics}). The plot illustrates that the Conditional Diffusion with Pose Estimation consistently achieves lower FID and higher MS-SSIM scores, indicating better visual quality and structural accuracy. In contrast, models without pose supervision tend to show variation, especially in complex or asymmetric postures. This highlights the importance of incorporating pose-level guidance to handle the anatomical and stylistic constraints of classical dance.
\begin{table}[!ht]
\centering
\small
\caption{Impact of Pose Estimation Modules on Synthesis Quality}
\begin{tabular}{|p{5cm}|c|c|}
\hline
\textbf{Model} & \textbf{FID ($\downarrow$)} & \textbf{MS-SSIM ($\uparrow$)}  \\ \hline
Conditional GAN (cGAN) & 26.27 & 0.71 \\ \hline
cGAN with Pose Estimation (CGPE) & 24.28 & 0.73 \\ \hline
Conditional Diffusion (cDiffusion)  & 21.88 & 0.75 \\ \hline
cDiffusion with Pose Estimation (CDMPE) & \textbf{19.32} & \textbf{0.78} \\ \hline
\end{tabular}
\label{tab:pose_module_impact}
\end{table}

\begin{table}[!ht]
\centering
\caption{Impact of Keypoint ($\mathcal{L}_\text{kp}$) and Pose consistency loss ($\mathcal{L}_\text{pose}$) on Synthesis Quality (ablation study)}
\begin{tabular}{|p{5cm}|c|c|}
\hline
\textbf{Model} & \textbf{FID ($\downarrow$)} & \textbf{MS-SSIM ($\uparrow$)} \\ \hline
cGAN with $\mathcal{L}_\text{kp}$ & 24.77 & 0.72 \\ \hline
cGAN with $\mathcal{L}_\text{pose}$ & 24.45 & 0.71 \\ \hline
cDiffusion with $\mathcal{L}_\text{kp}$ & 20.15 & 0.76 \\ \hline
cDiffusion with $\mathcal{L}_\text{pose}$ & 20.11 & 0.77 \\ \hline
CDMPE (with $\mathcal{L}_\text{kp}$ and $\mathcal{L}_\text{pose}$) & \textbf{19.32} & \textbf{0.78} \\ \hline
\end{tabular}
\label{tab:loss_impact}
\end{table}

\subsection{Comparative Analysis}
To the best of our knowledge, no prior work explicitly addresses the task of generating key postures for classical dance forms. Therefore, we benchmark our method against existing full-body human image generation baselines, which predominantly focus on unconditional generation. Refer Table~\ref{image_synthesis_comparison} for Quantitative comparison of key posture generation performance using FID and MS-SSIM metrics. Our method achieves an FID of 19.32 and an MS-SSIM of 0.78, indicating competitive performance with enhanced structural consistency. Our method demonstrates superior qualitative alignment with culturally and anatomically accurate pose structures. This alignment is critical in the context of structured dance generation, where stylistic precision outweighs diversity.

\subsection{Qualitative Analysis and Observations}

We present selected samples of generated {\em Adavu} key postures alongside real (ground truth) images in Figure~\ref{result} to qualitatively assess the performance of our models. These comparisons illustrate the capacity of each method to generate culturally grounded and pose-consistent {\em Bharatanatyam} movements. Notably, Figure~\ref{best_result} highlights the outputs of our Conditional Diffusion model with pose estimation, which demonstrates superior preservation of structural accuracy and stylistic fidelity. A key advantage of our approach lies in the integration of a pose estimation module that enforces precise geometric alignment and anatomical correctness. Unlike generative models that may introduce stylistic variations—such as exaggerated gestures or fluidity for visual appeal—our method adheres strictly to the codified pose structure of classical dance. This fidelity is critical in the context of Intangible Cultural Dances (ICDs), where each pose is imbued with symbolic, historical, and often spiritual significance.By constraining the generation to structurally accurate outputs, the model ensures cultural respect and authenticity. Moreover, this methodology is generalizable to other rule-based or posture-sensitive dance forms, including Indian classical styles, martial ritual performances, and codified folk traditions, where precise limb coordination and timing are essential for education, preservation, and respectful representation.

\begin{table}[ht]
\small
\centering
\caption{Quantitative comparison of key posture generation performance}
\begin{tabular}{|p{5cm}|c|c|}
\hline
\textbf{Method} & \textbf{FID ($\downarrow$)} & \textbf{MS-SSIM ($\uparrow$)} \\
\hline
InsetGAN \cite{Fruehstueck2022InsetGAN}& 20.72 & 0.77 \\ \hline
PCGAN \cite{Liang2019PCGAN} & 21.17 & 0.75 \\ \hline
XingGAN  \cite{tang2020xinggan} & 22.43 & 0.76\\ \hline
StyleGAN-Human  \cite{Fu2022StyleGANHuman} & 20.32 & 0.74 \\ \hline
PIDM  \cite{bhunia2022person}& 23.57 & 0.71 \\ \hline
\textbf{CDMPE (Our Method)} & \textbf{19.32} & \textbf{0.78} \\
\hline
\end{tabular}
\label{image_synthesis_comparison}
\end{table}

\section{Conclusion}
\label{sec;conclusion}
Generating key-postures of {\em Bharatanatyam} dance through generative models offers a novel approach to capturing its essence, with these postures forming the foundation for synthesizing Adavu videos. Our methodology utilizes different generative models, incorporating pose estimation further ensures alignment with predefined body postures and allows key point-based feedback, crucial for maintaining the geometric and rhythmic precision, to effectively generate realistic and rule-compliant dance postures central to {\em Bharatanatyam}. These approaches ensure that intricate movements, codified gestures, and stylistic nuances of {\em Indian Classical Dance} are captured with precision and fidelity to cultural norms. Despite advancements, challenges remain in replicating facial expressions essential to {\em Bharatanatyam}’s {\em abhinaya}, as models struggle with nuanced emotions and eye movements. Addressing these challenges is essential for improving the fidelity and expressiveness of generated dance poses. Despite these hurdles, the integration of AI-driven models in {\em Bharatanatyam} has the potential to revolutionize the preservation and dissemination of this cultural heritage, ensuring its relevance and appreciation in the digital and global era. Our framework is generalizable to other codified dance or ritual movement systems that require precise pose representation, offering utility for digital preservation, pedagogy, and avatar animation in culturally sensitive domains.

\bibliographystyle{ACM-Reference-Format}
\bibliography{ICVGIP-Latex-Template}

@String{Computing = "Computing" }

@String{Computer = "{IEEE} Computer" }

@String{Springer = "Springer-Verlag" }

@article{kalpana2015bharatanatyam,
  title={Bharatanatyam and Mathematics: Teaching Geometry Through Dance},
  author={Kalpana, Iyengar Mukunda},
  journal={Journal of Fine and Studio Art},
  volume={5},
  number={2},
  pages={6--17},
  year={2015},
  doi={10.5897/JFSA2015.0031},
}

@misc{ranganathan_design_bharatanatyam,
author = {Ranganathan, Jaikumar},
year = {2019},
month = {05},
pages = {},
title = {Decoding the Principles of Design and Sacred Geometry},
doi = {10.13140/RG.2.2.20749.38887}
}

@inproceedings{li2021learn,
  title={Learn to Dance with AIST++: Music Conditioned 3D Dance Generation},
  author={Li, Ruilong and Yang, Shan and Ross, David A. and Kanazawa, Angjoo},
  booktitle={International Conference on Computer Vision (ICCV)},
  pages={13401--13411},
  year={2021},
  organization={IEEE}
}

@misc{NTU_RGBD,
  author  = {Shahroudy, Amir and Liu, Jun and Ng, Tian-Tsong and Wang, Gang},
  title   = {NTU RGB+D Action Recognition Dataset},
  year    = {2016},
}

@inbook{bhuyan2021keyframes,
author = {Bhuyan, Himadri and Das, Parthapratim},
year = {2021},
month = {03},
pages = {174-185},
title = {Recognition of Adavus in Bharatanatyam Dance},
publisher = {Springer, Singapore},
isbn = {978-981-16-1102-5},
doi = {10.1007/978-981-16-1103-2_16}
}

@inproceedings{kaushik2018nrityantar,
  author    = {Vinay Kaushik, Prerana , Brejesh Lall},
  title     = {Nrityantar: Pose Oblivious Indian Classical Dance Sequence Classification System},
  booktitle = {Indian Conference on Computer Vision, Graphics and Image Processing},
  year      = {2018},
}

@inproceedings{luo2024popdg,
  author    = {Zhenye Luo and Min Ren and Xuecai Hu and Yongzhen Huang and Li Yao},
  title     = {POPDG: Popular 3D Dance Generation with PopDanceSet},
  booktitle = {Proceedings of the IEEE/CVF Conference on Computer Vision and Pattern Recognition (CVPR)},
  year      = {2024},
  pages     = {26984--26993},
  doi       = {10.1109/CVPR52688.2024.02693},
}

@article{wang2024musicdrivendance,
  author    = {Wang, Huaxin and Song, Yang and Jiang, Wei and Wang, Tianhao},
  title     = {A Music-Driven Dance Generation Method Based on a Spatial-Temporal Refinement Model to Optimize Abnormal Frames},
  journal   = {Sensors},
  year      = {2024},
  volume    = {24},
  number    = {2},
  pages     = {588},
  doi       = {10.3390/s24020588},
}

@inproceedings{chan2019everybody,
  author    = {Caroline Chan and Shiry Ginosar and  Tinghui Zhou and Alexei Efros},
  title     = {Everybody dance now},
  booktitle = {International Conference on Computer Vision},
  year      = {2019},
  pages     = {5933--5942},
}

@inproceedings{seong2021keypoint,
  title={Keypoint-wise Adaptive Loss for Whole-Body Human Pose Estimation},
  author={Bongjo Seong and Hyug-Jae Lee and Rokkyu Lee},
  year={2023},
  url={https://api.semanticscholar.org/CorpusID:263630221}
}

@article{raju2024pose2gest,
  author = {Kavitha Raju and Nandini J. Warrier and Manu Madhavan, Selvi C. and Arun B. Warrier and Thulasi Kumar},
  title = {Pose2Gest: A Few-Shot Model-Free Approach Applied In South Indian Classical Dance Gesture Recognition},
  year = {2024},
  journal = {arXiv},
  volume = {abs/2404.11205},
}

@article{jain2021enhanced,
  author    = {Jain, N. and Bansal, V. and Virmani, D. and Gupta, V. and Salas-Morera, L. and Garcia-Hernandez, L.},
  title     = {An Enhanced Deep Convolutional Neural Network for Classifying Indian Classical Dance Forms},
  journal   = {Applied Sciences},
  year      = {2021},
  volume    = {11},
  pages     = {6253},
  doi       = {10.3390/app11146253},
}

@article{mallick2019posture,
  title={Posture and Sequence Recognition for Bharatanatyam Dance Performances Using Machine Learning Approach},
  author={Mallick, Tanwi and Das, Partha Pratim and Majumdar, Arun Kumar},
  journal={arXiv preprint arXiv:1909.11023},
  year={2019},
  url={https://arxiv.org/abs/1909.11023}
}

@article{mirza2014conditional,
  author    = {Mehdi Mirza and Simon Osindero},
  title     = {Conditional Generative Adversarial Nets},
  journal   = {arXiv preprint arXiv:1411.1784},
  year      = {2014},
  doi       = {10.48550/arXiv.1411.1784},
}

@inproceedings{ho2020denoising,
  title={Denoising Diffusion Probabilistic Models},
  author={Ho, Jonathan and Jain, Ajay and Abbeel, Pieter},
  booktitle={Advances in Neural Information Processing Systems},
  volume={33},
  pages={6840--6851},
  year={2020},
  organization={Curran Associates, Inc.}
}

@article{ho2022classifier,
  title={Classifier-Free Diffusion Guidance},
  author={Ho, Jonathan and Salimans, Tim},
  journal={arXiv preprint arXiv:2207.12598},
  year={2022}
}

@inproceedings{lokumannage2021bharatanatyam,
  author    = {Lokumannage, A.},
  title     = {A Study of Non-Verbal Communication Symbolic Meanings Reflected in Bharatanatyam Gestures},
  booktitle = {Proceedings of the Open University Research Sessions (OURS 2021)},
  year      = {2021}
}

@article{chakraborty2019dance,
  title={Dance (Bharatanatyam): The Art of Non-Verbal Communication},
  author={Chakraborty, Aishika and Moin, Syed Wasif and Dey, Arpita and Bose, Ankita},
  journal={International Journal of English Learning and Teaching Skills},
  volume={1},
  number={3},
  pages={251--258},
  year={2019},
  publisher={Smart Society}
}

@misc{mediapipe2020,
  author = {Google},
  title = {MediaPipe: A Framework for Building Perception Pipelines},
  year = {2020},
  url = {https://mediapipe.dev/},
}

@inproceedings{heusel2017gans,
  author    = {Martin Heusel and Hubert Ramsauer and Thomas Unterthiner and Bernhard Nessler and Sepp Hochreiter},
  title     = {GANs Trained by a Two Time-Scale Update Rule Converge to a Local Nash Equilibrium},
  booktitle = {Advances in Neural Information Processing Systems (NeurIPS)},
  year      = {2017},
  pages     = {6626--6637},
  archivePrefix = {arXiv},
  eprint    = {1706.08500},
}

@misc{jayasumana2024rethinkingfidbetterevaluation,
      title={Rethinking FID: Towards a Better Evaluation Metric for Image Generation}, 
      author={Sadeep Jayasumana and Srikumar Ramalingam and Andreas Veit and Daniel Glasner and Ayan Chakrabarti and Sanjiv Kumar},
      year={2024},
      eprint={2401.09603},
      archivePrefix={arXiv},
      primaryClass={cs.CV},
}

@article{liu2022improved,
  title={Improved Fine-Tuning by Better Leveraging Pre-Training Data},
  author={Liu, Ziquan and Xu, Yi and Xu, Yuanhong and Qian, Qi and Li, Hao and Ji, Xiangyang and Chan, Antoni B. and Jin, Rong},
  journal={arXiv:2111.12292v2},
  year={2022},
}

@inproceedings{wang2003multiscale,
  author    = {Zhou Wang and Eero P. Simoncelli and Alan C. Bovik},
  title     = {Multiscale Structural Similarity for Image Quality Assessment},
  booktitle = {The 37th Asilomar Conference on Signals, Systems and Computers},
  year      = {2003},
  pages     = {1398-1402},
  doi       = {10.1109/ACSSC.2003.1292216}
}

@inproceedings{lucic2018gans,
  title={Are GANs Created Equal? A Large-Scale Study},
  author={Lucic, Mario and Kurach, Karol and Michalski, Marcin and Gelly, Sylvain and Bousquet, Olivier},
  booktitle={Advances in Neural Information Processing Systems (NeurIPS)},
  year={2018},
  pages={700--709}
}

@inproceedings{sajjadi2018assessing,
  title={Assessing Generative Models via Precision and Recall},
  author={Sajjadi, Mehdi S. M. and Bachem, Olivier and Lucic, Mario and Bousquet, Olivier and Gelly, Sylvain},
  booktitle={Advances in Neural Information Processing Systems (NeurIPS)},
  year={2018},
  pages={5228--5237}
}

@article{borji2019pros,
  title={Pros and Cons of GAN Evaluation Measures},
  author={Borji, Ali},
  journal={Computer Vision and Image Understanding},
  volume={179},
  pages={41--65},
  year={2019},
  publisher={Elsevier}
}

@article{kishore2018indian,
  title     = {Indian Classical Dance Action Identification and Classification with Convolutional Neural Networks},
  author    = {Kishore, P. V. V. and Kumar, K. V. K. Vijay and Eepuri, Kiran Kumar and Prasad, M. V. D.},
  journal   = {Advances in Multimedia},
  volume    = {2018},
  pages     = {1--17},
  year      = {2018},
  publisher = {Hindawi},
  doi       = {10.1155/2018/5141402},
}

@article{kumar2017indian,
  title     = {Indian Classical Dance Classification with Adaboost Multiclass Classifier on Multifeature Fusion},
  author    = {Kumar, K. V. K. Vijay and Kishore, P. V. V. and Kumar, D. Anil},
  journal   = {Mathematical Problems in Engineering},
  volume    = {2017},
  pages     = {1--18},
  year      = {2017},
  publisher = {Hindawi},
  doi       = {10.1155/2017/6204742},
}

@article{jayanthi2024ai,
  title     = {AI and Augmented Reality for 3D Indian Dance Pose Reconstruction: Cultural Revival},
  author    = {Jayanthi, J. and Uma Maheswari, P.},
  journal   = {Heliyon},
  year      = {2024},
  publisher = {Elsevier},
  pmid      = {38575710},
  pmcid     = {PMC10994917},
  doi       = {10.1016/j.heliyon.2024.e26439},
}

@article{sharma2013bharatanatyam,
  title     = {Bharatanatyam: The Crescendo of Non-Verbal Communication},
  author    = {Sharma, Preeti Bala},
  journal   = {IMPACT: International Journal of Research in Humanities, Arts and Literature (IMPACT: IJRHAL)},
  volume    = {1},
  number    = {5},
  pages     = {13--22},
  year      = {2013}
}

@inproceedings{bhunia2022person,
  title={Person Image Synthesis via Denoising Diffusion Model},
  author={Bhunia, Ankan Kumar and Khan, Salman and Cholakkal, Hisham and Anwer, Rao Muhammad and Laaksonen, Jorma and Shah, Mubarak and Khan, Fahad Shahbaz},
  booktitle={Proceedings of the IEEE/CVF Conference on Computer Vision and Pattern Recognition (CVPR)},
  year={2022},
  pages={10600--10610}
}

@article{Morris01042003,
author = {Geraldine Morris and},
title = {Problems with Ballet: Steps, style and training},
journal = {Research in Dance Education},
volume = {4},
number = {1},
pages = {17--30},
year = {2003},
publisher = {Routledge},
doi = {10.1080/14647890308308},
}

@online{unesco,
  author       = {{UNESCO}},
  title        = {Intangible Cultural Heritage},
  year         = {2024},
  url          = {https://ich.unesco.org/en/home},
  note         = {Accessed: 2024-06-20}
}

@inproceedings{Fruehstueck2022InsetGAN,
  author    = {Anna Fr{\"u}hst{\"u}ck and Krishna Kumar Singh and Eli Shechtman and Niloy J. Mitra and Peter Wonka and Jingwan Lu},
  title     = {InsetGAN for Full-Body Image Generation},
  booktitle = {Proceedings of the IEEE/CVF Conference on Computer Vision and Pattern Recognition (CVPR)},
  year      = {2022},
  pages     = {7723--7732},
  publisher = {IEEE},
  address   = {New Orleans, LA, USA},
  month     = {June}
}

@inproceedings{Liang2019PCGAN,
  author       = {Dong Liang and Rui Wang and Xiaowei Tian and Cong Zou},
  title        = {PCGAN: Partition-Controlled Human Image Generation},
  booktitle    = {Proceedings of the Thirty-Third AAAI Conference on Artificial Intelligence (AAAI-19)},
  year         = {2019},
  pages        = {7959--7966},
  publisher    = {AAAI Press},
  address      = {Honolulu, HI, USA},
  month        = {January}
}

@inproceedings{tang2020xinggan,
  title={XingGAN for Person Image Generation},
  author={Tang, Hao and Bai, Song and Zhang, Li and Torr, Philip H.S. and Sebe, Nicu},
  booktitle={European Conference on Computer Vision (ECCV)},
  year={2020},
  pages={85--102},
  publisher={Springer},
  doi={10.1007/978-3-030-58571-6_6}
}

@inproceedings{Fu2022StyleGANHuman,
  author       = {Jianglin Fu and Shikai Li and Yuming Jiang and Kwan-Yee Lin and Chen Qian and Chen Change Loy and Wayne Wu and Ziwei Liu},
  title        = {StyleGAN-Human: A Data-Centric Odyssey of Human Generation},
  booktitle    = {Proceedings of the European Conference on Computer Vision (ECCV)},
  year         = {2022},
  doi          = {10.1007/978-3-031-19787-1_1},
  publisher    = {Springer},
  address      = {Glasgow, UK},
  month        = {August}
}

@article{deBem2020DGPose,
  author    = {de Bem, Rodrigo and Ghosh, Arnab and Ajanthan, Thalaiyasingam and Miksik, Ondrej and Boukhayma, Adnane and Siddharth, N. and Torr, Philip H.S.},
  title     = {DGPose: Deep Generative Models for Human Body Analysis},
  journal   = {International Journal of Computer Vision},
  volume    = {128},
  pages     = {1537--1563},
  year      = {2020},
  doi       = {10.1007/s11263-020-01306-1},
}

@article{Xu2023PACGAN,
  author    = {Xu, Cheng and Chen, Zejun and Mai, Jiajie and Xu, Xuemiao and He, Shengfeng},
  title     = {Pose- and Attribute-consistent Person Image Synthesis},
  journal   = {ACM Transactions on Multimedia Computing, Communications and Applications},
  volume    = {19},
  number    = {2},
  pages     = {1--21},
  year      = {2023},
  doi       = {10.1145/3554739},
}

@article{Zhang2024MimicMotion,
  title       = {MimicMotion: High-Quality Human Motion Video Generation with Confidence-aware Pose Guidance},
  author      = {Zhang, Yuang and Gu, Jiaxi and Wang, Li-Wen and Wang, Han and Cheng, Junqi and Zhu, Yuefeng and Zou, Fangyuan},
  journal     = {arXiv preprint arXiv:2406.19680},
  year        = {2024},
  month       = jun,
}

@online{GeometryInMotion2021,
  author    = {Balaji, Vajra},
  title     = {Geometry in Motion: The Symmetry of Bharatanatyam Poses},
  year      = {2021},
  url       = {https://www.indian-dance.com/post/geometry-in-motion-the-symmetry-of-bharatanatyam-poses},
  note      = {Accessed: 2023-12-02}
}

@online{BalletBasics2015,
  author    = {{Sarah}},
  title     = {Ballet Basics: Posture and Alignment},
  year      = {2015},
  month     = oct,
  day       = 27,
  url       = {https://learnaboutballet.wordpress.com/2015/10/27/ballet-basics-posture-and-alignment/},
  note      = {Accessed: 2024-03-2}
}

@misc{Mallick2017KathakGeometry,
  author       = {Mallick, Sandip},
  title        = {The Geometric Analysis with a Chronological Order for the Basic Movements and Footwork used in Kathak},
  year         = {2017},
  howpublished = {Junior Fellowship Project Report, Centre for Cultural Resources \& Training (CCRT)},
  
}

@article{Daprati2009DanceToTheMusic,
  author    = {Daprati, Elena and Iosa, Marco and Haggard, Patrick},
  title     = {A Dance to the Music of Time: Aesthetically-Relevant Changes in Body Posture in Performing Art},
  journal   = {PLoS ONE},
  volume    = {4},
  number    = {3},
  pages     = {e5023},
  year      = {2009},
  month     = mar,
  doi       = {10.1371/journal.pone.0005023},
}

@incollection{Dhananjayan2002Transmission,
  author    = {Dhananjayan, V.P.},
  title     = {Beyond Performing: The Bharatanatyam Dancer’s Role in Cultural Transmission},
  booktitle = {Natya Kala Conference Proceedings},
  year      = {2002},
  publisher = {Krishna Gana Sabha},
}

@book{Vatsyayan1997Theatre,
  author    = {Vatsyayan, Kapila},
  title     = {Traditional Indian Theatre: Multiple Streams},
  year      = {1997},
  publisher = {IGNCA and Sangeet Natak Akademi},
  address   = {New Delhi},
 
}

@misc{ChallengesClassical2016,
  author       = {Ilmi, Aijaz},
  title        = {Challenges for classical Indian dance},
  year         = {2016},
  howpublished = {The Asian Age (online)},
}

@article{ClassicalGatekeeping2021,
  author       = {Basu, Priyanka},
  title        = {Restitution and the Writing of Indian 'Classical' Dance: Rethinking Social Justice for Marginal Communities},
  journal      = {Roots§Routes Journal},
  year         = {2021},
}

@inproceedings{ma2017pose,
  title={Pose guided person image generation},
  author={Ma, Liqian and Jia, Xu and Sun, Qianru and Schiele, Bernt and Tuytelaars, Tinne and Van Gool, Luc},
  booktitle={NeurIPS},
  year={2017}
}

@inproceedings{ma2018disentangled,
  title={Disentangled person image generation},
  author={Ma, Liqian and Sun, Qianru and Georgoulis, Stamatios and Van Gool, Luc and Schiele, Bernt and Fritz, Mario},
  booktitle={CVPR},
  year={2018},
  pages={99--108}
}

@inproceedings{song2019unsupervised,
  title={Unsupervised person image generation with semantic parsing transformation},
  author={Song, Sijie and Zhang, Wei and Liu, Jiaying and Mei, Tao},
  booktitle={CVPR},
  year={2019},
  pages={2357--2366}
}

@inproceedings{gong2017look,
  title={Look into person: Self-supervised structure-sensitive learning and a new benchmark for human parsing},
  author={Gong, Ke and Liang, Xiaodan and Zhang, Dongyu and Shen, Xiaohui and Lin, Liang},
  booktitle={CVPR},
  year={2017},
  pages={932--940}
}

@inproceedings{liu2016deepfashion,
  title={Deepfashion: Powering robust clothes recognition and retrieval with rich annotations},
  author={Liu, Ziwei and Luo, Ping and Qiu, Shi and Wang, Xiaogang and Tang, Xiaoou},
  booktitle={CVPR},
  year={2016},
  pages={1096--1104}
}

@inproceedings{jetchev2017conditional,
  title={The conditional analogy GAN: Swapping fashion articles on people images},
  author={Jetchev, Nikolay and Bergmann, Urs},
  booktitle={ICCV Workshops},
  year={2017},
  pages={2287--2292}
}

@inproceedings{cao2017realtime,
  title={Realtime multi-person 2D pose estimation using part affinity fields},
  author={Cao, Zhe and Simon, Tomas and Wei, Shih-En and Sheikh, Yaser},
  booktitle={CVPR},
  year={2017},
  pages={7291--7299}
}

@inproceedings{balakrishnan2018synthesizing,
  title={Synthesizing images of humans in unseen poses},
  author={Balakrishnan, Guha and Zhao, Amy and Dalca, Adrian V and Durand, Fredo and Guttag, John},
  booktitle={CVPR},
  year={2018},
  pages={8340--8348}
}

@inproceedings{goodfellow2014generative,
  title={Generative adversarial nets},
  author={Goodfellow, Ian and Pouget-Abadie, Jean and Mirza, Mehdi and Xu, Bing and Warde-Farley, David and Ozair, Sherjil and Courville, Aaron and Bengio, Yoshua},
  booktitle={Advances in neural information processing systems},
  volume={27},
  year={2014}
}

@misc{lee2023embedding,
  title={Embedding for Pose-Guided Person Image Synthesis with Diffusion Model},
  author={Donghwan Lee and Kyungha Min and Kirok Kim and Seyoung Jeong and Jiwoo Jeong and Wooju Kim},
  year={2023},
  note={Manuscript, Yonsei University},
  url={https://arxiv.org/abs/2412.07333} 
}

@inproceedings{zhao2022stylegan,
  title={StyleGAN-Human: A Data-Centric Odyssey of Human Generation},
  author={Zhao, Yujun and Zhang, Yifan and He, Haonan and Zhu, Tianjia and Sun, Qianru and Loy, Chen Change},
  booktitle={Proceedings of the European Conference on Computer Vision (ECCV)},
  pages={1--19},
  year={2022}
}

\pagebreak
\appendix

\end{document}